# Brain Intelligence: Go Beyond Artificial Intelligence

Huimin Lu[1,*], Yujie Li[2], Min Chen[3], Hyoungseop Kim[1], Seiichi Serikawa[1]

[1]Kyushu Institute of Technology, Japan

[2]Yangzhou University, China

[3]Huazhong University of Science and Technology, China

luhuimin@ieee.org

**Abstract**

Artificial intelligence (AI) is an important technology that supports daily social life and economic activities. It contributes greatly to the sustainable growth of Japan's economy and solves various social problems. In recent years, AI has attracted attention as a key for growth in developed countries such as Europe and the United States and developing countries such as China and India. The attention has been focused mainly on developing new artificial intelligence information communication technology (ICT) and robot technology (RT). Although recently developed AI technology certainly excels in extracting certain patterns, there are many limitations. Most ICT models are overly dependent on big data, lack a self-idea function, and are complicated. In this paper, rather than merely developing next-generation artificial intelligence technology, we aim to develop a new concept of general-purpose intelligence cognition technology called "Beyond AI". Specifically, we plan to develop an intelligent learning model called "Brain Intelligence (BI)" that generates new ideas about events without having experienced them by using artificial life with an imagine function. We will also conduct demonstrations of the developed BI intelligence learning model on automatic driving, precision medical care, and industrial robots.

## 1. Introduction

From SIRI [1] to AlphaGo [2], artificial intelligence (AI) is developing rapidly. While science fiction often portrays AI as robots with human-like characteristics, AI can encompass anything from e-Commerce prediction algorithms to IBM's Watson machines [3]. However, artificial intelligence today is properly known as weak AI, which is designed to perform a special task (e.g., only facial recognition or only internet searches or only driving a car). While weak AI may outperform humans at a specific task, such as playing chess or solving equations, general AI would outperform humans at nearly every cognitive task.

In recent years, the US government has supported basic research on AI, which is centered on robots and pattern recognition (voice, images, etc.). Microsoft has announced real-time translation robots and innovative image recognition technologies [4]. Amazon uses artificial intelligence for autonomous robots in delivery systems [5]. Facebook has also developed facial recognition technology based on artificial intelligence called "DeepFace" [6]. Robots and artificial intelligence are being actively studied in university institutions in the United States. Innovative technologies, such as corporate cooperation and deep learning, are emerging. The robot car developed by the Artificial Intelligence Laboratory at Stanford University has set a faster time than an active racer [7]. The Computer Science and Artificial Intelligence Laboratory at Massachusetts Institute of Technology has developed a cleaning robot and a four-foot walking robot [8].

Meanwhile, AI is the main technology expected to improve Japanese ICT's innovation and robot technology in the near future. ICT in Japan has rapidly advanced in recent years. To secure Japan's status as a world-class "technological superpower", the Japanese government has formulated projects such as the "Science and Technology Basic Plan [9]" and the "Artificial Intelligence Technology Conference [10]". Japan is expecting to utilize state-of-the-art artificial intelligence and robots to solve various problems.

However, through some research, we found that recent artificial intelligence technologies have many limitations. In the following, we list some representative limitations and analyze the reasons why recent AI cannot break through these inherent disadvantages.

**Limitations of Artificial Intelligence**

In recent years, artificial intelligence technologies have developed dramatically due to improvement in the processing capacity of computers and the accumulation of big data. However, the results of current artificial intelligence technologies remain limited to specific intellectual areas, such as image recognition, speech recognition, and dialogue response. That is, current AI is a specialized type of artificial intelligence acting intellectually in a so-called individual area (see Figure 1). Examples include techniques such as Convolutional Neural Networks (CNN) or Deep Residual Learning (ResNet) for visual recognition, Recurrent Neural Networks (RNN) or Deep Neural Networks (DNN) for speech recognition, and Represent Learning (RL) for dialogue understanding. All of these are a part of the intellectual work carried out by each area of the human brain; they are only a substitute and do not perform all of the functions of the human brain. In other words, AI has not been able to cooperate with whole-brain functions such as self-understanding, self-control, self-consciousness and self-motivation. Specifically, we conclude that the limitations of the recent artificial intelligence technologies are the following:

(1) *Frame Problem*

Considering all the events that can occur in the real world, since it takes a large amount of time due to big data training, AI is typically limited to a single frame or type of problem. For example, if you restrict the algorithm to apply only to chess, shogi, image recognition, or speech recognition, only certain results can be expected. However, when trying to cope with every phenomenon in the real world, there is an infinite number of possibilities that we have to anticipate, so the extraction time becomes infinite due to overloading of the database.

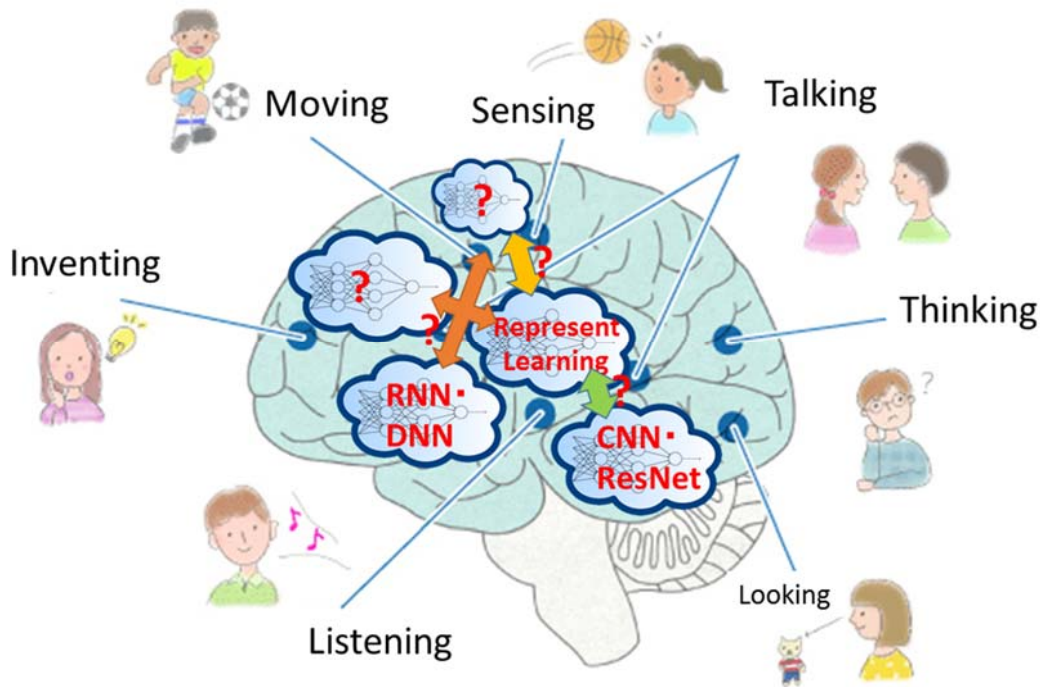

Figure 1. Shortages of current artificial intelligence.

(2) *Association Function Problem*

Machine learning and artificial intelligence are excellent at extracting a particular pattern. However, the results of machine learning are easy to misuse. Current artificial intelligence technology depends on large-scale data and can obtain results using only numerical values, but it does not have the association function like the human brain. That is, a single part of the brain cannot be as intelligent as the whole brain.

(3) *Symbol Grounding Problem*

It is necessary to link symbols with their meanings, but this task is often not resolved in current artificial intelligence. For example, if you know the individual meaning of the word "horse" and the meaning of the word "stripes", then when you are taught that "zebra = horse + stripes", you can understand that "a zebra is a horse with stripes". However, the computer cannot make the same connections between ideas.

(4) *Mental and Physical Problem*

What is the relationship between the mind and body? That is, if the mind is generally thought of as non-material, how can the physical body be affected by it? Whether or not this is possible has not been elucidated.

In conclusion, we can see that there are many problems unsolved in current artificial intelligence. In this paper, we first review the most recent algorithms for weak AI. Then, we introduce the next-generation intelligence architecture, Brain Intelligence, which is an advanced artificial intelligence for solving the disadvantages of weak AI algorithms.

## 2. Artificial Intelligence

The market and business for AI technologies is changing rapidly. In addition to speculation and increased media attention, many start-up companies and Internet giants are racing to acquire AI technologies in business investment. Narrative Science Survey found that 38% of enterprises have been using AI in 2016, and the number will increase to 62% in 2018. Forrester Research expects AI investment in 2017 to grow by more than 300% compared with 2016. IDC estimates that the AI market will grow from $ 8 billion in 2016 to $ 47 billion in 2020 [11].

Current artificial intelligence includes a variety of technologies and tools, some time-tested and others that are relatively new. To help understand what is hot and what is not, Forrester has just released a TechRadar report on artificial intelligence (application developers), detailing the 9 technologies for which companies should consider using artificial intelligence to support decisions.

### 2.1 Natural Language Generation

Natural language generation (NLG) is used to generate text from computer data using AI, especially deep learning architectures, to generate the NLG tasks. Deep neural networks (DNN) are undoubtedly one of the most popular research areas in the current NLG field. DNN are designed to learn representations at increasing layers of abstraction by adopting backpropagation [12], feedforwards [13], log-bilinear models [14], and recurrent neural networks (RNN) [15]. Their advantage over traditional models is that DNN models represent voice sequences of varying lengths, so similar histories have related representations. They overcome the disadvantage of traditional models, which have data sparseness and a recorder for remembering the parameters.

Long short-term memory architectures (LSTM) [16], which are a further development upon RNN, contain memory cells and multiplicative gates that control the information's access. Mei et al. [17] proposed a LSTM-based architecture, which uses the encoder-decoder framework, for content selection and realization. Luong et al. [18] showed that parsing the datasets used for co-training in encoder and decoder can improve the translation efficiency. In most of these methods, it is hard to balance between achieving adequate textual output and generating text efficiently and robustly.

LTSM is currently used for customer service, report generation and summary of business intelligence insight. Examples of suppliers include Attivio, Automated Insights, Cambridge Semantics, Digital Reasoning, Lucidworks, Narrative Science, SAS, and Yseop.

### 2.2 Speech Recognition

Hidden Markov models (HMM) [19] are useful tools for speech recognition. In recent years, deep feedforward networks (DFN) have gained attention for solving issues of recognition. It seems that

HMM combines with RNN as a better solution. However, the HMM-RNN model does not perform as well as deep networks. Speech recognition's goal is to translate human language and convert it into a useful format for computer applications. Graves et al. [20] proposed a deep long short-term memory RNNs for speech recognition. This model is an end-to-end learning method that jointly trains two separate RNNs as acoustic and linguistic models. It is widely used in current interactive voice response systems and mobile applications. Examples of suppliers include NICE, Nuance Communications, OpenText, and Verint Systems.

**2.3 Virtual/Augmented Reality**

Virtual reality uses simple devices and advanced systems that can network with humans. Virtual reality is a computer-generated simulation of a 3D environment that can be interacted with in a seemingly real manner. Artificial intelligence will be used in augmented reality for future remote eHealth [21, 22]. It is currently used in customer service and support and as a smart home manager [23, 49-52]. Sample vendors include Amazon, Apple, Artificial Solutions, Assist AI, Creative Virtual, Google, IBM, IPsoft, Microsoft, and Satisfi.

**2.4 AI-optimized Hardware**

Because of the rapid growth of data in recent years, it is possible for engineers to use the massive amounts of data to learn patterns. Most artificial intelligence models are proposed to meet these needs. These models require a large amount of data and computing power to train and are limited by the need for better hardware acceleration to accommodate scaling beyond current data and model sizes. Graphics processing units (GPU) [24], general purpose processors (GPGPU) [25] and field programmable gate arrays (FPGA) [26] are required to efficiently run AI-oriented computational tasks. GPU has orders of magnitude more computational cores than traditional general purpose processors (GPP) and allows a greater ability to perform parallel computations. In particular, GPGPU is usually used. Unlike GPUs, FPGA has a flexible hardware configuration, and provides better performance per Watt than GPUs. However, it is difficult to program FPGA devices because of the special architecture. Sample vendors include Alluviate, Cray, Google, IBM, Intel, and Nvidia.

**2.5 Decision Management**

Decision-making plays a critical role in achieving sustainable development during turbulent financial markets. With the improvement of information communication technology (ICT), AI-based techniques, such as decision tree (DT), support vector machine (SVM), neural network (NN), and deep learning, have been used for decision making [27]. Engines that insert rules and logic into AI systems are used for initial setup/training and ongoing maintenance and tuning. A mature technology, which is used in a wide variety of enterprise applications, assisting in or performing automated decision making. Sample vendors include Advanced Systems Concepts, Informatica, Maana, Pegasystems, and UiPath.

**2.6 Deep Learning Platforms**

Currently, research used in pattern recognition and classification is primarily supported by very large data sets. Few approaches look for providing a solution better than existing big-data processing platforms, which usually runs over a large-scale commodity CPU cluster. Moreover, GPUs seem to the best platforms to train AI networks [28]. However, recent platforms are worse than the human brain in processing perception and require large amounts of space and energy. To this end, Rajat et al. [29] trained a DBN model with 100 million parameters using an Nvidia GTX280 graphics card with 240 cores. Adam et al. [30] proposed a COTS HPC deep neural network training system. Google developed DistBelief [31], which uses thousands of CPUs to train the deep neural network; see Figure 2. Microsoft developed project Adam [32] to use fewer machines for training. Other sample vendors, such as Qualcomm's Zeroth platform [33], IBM's Truenorth [34], and Manchester University's SpiNNaker [35] are also in development. Besides, there are also some software packages for deep learning. These packages include Tensorflow, Theano, Torch/PyTorch, MxNet, Caffe as well as high level package Keras. It would also be good to mention Google's TPU when mentioning the hardware platforms.

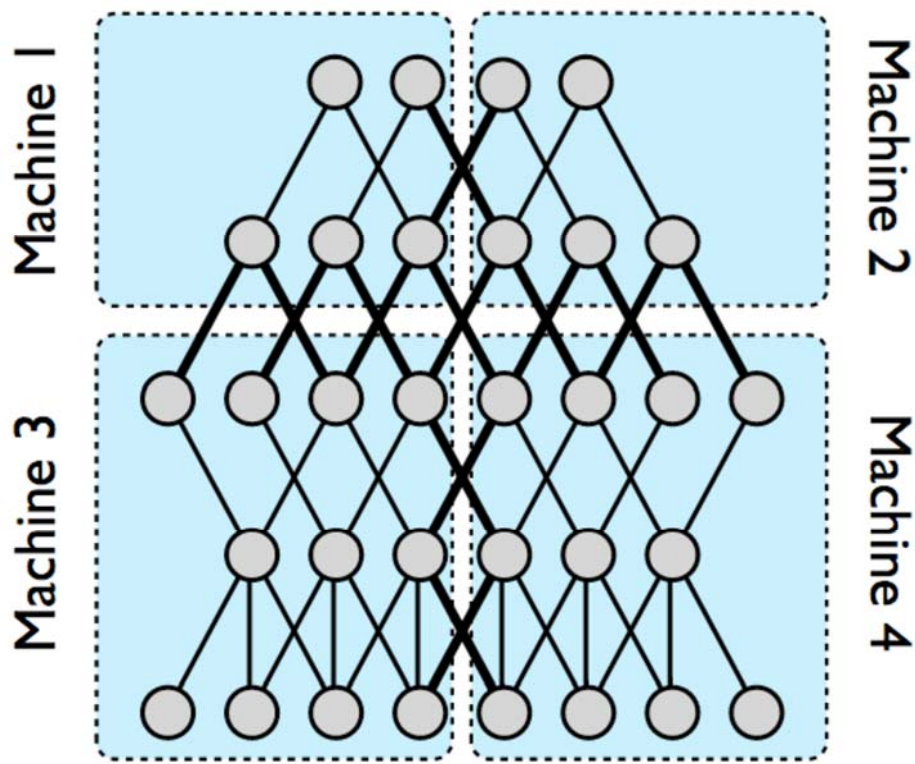

Figure 2 Google DistBelief approach for DNN training [31].

## 2.7 Robotic Process Automation

Robotic process automation (RPA) [36] uses software and algorithms to automate human action to support efficient business processes. A software robot is used instead of humans for typing and clicking and even analyzing data in different applications. RPA is currently used where it is too expensive or inefficient for humans to execute a task or a process. Researchers are promoting the adoption of RPA in the financial area. RPA has also been applied to trading in treasuries, affecting accounting staff involved in the banking area. AI, as a solution for big data, provides a new possibility for accurate prediction of RPA. Sample vendors include Advanced Systems Concepts, Automation Anywhere, Blue Prism, UiPath, and WorkFusion.

## 2.8 Text Analytics and NLP

Natural language processing (NLP) uses and supports text analytics by facilitating the understanding of sentence structure and meaning, sentiment, and intent through statistical and machine learning methods. NLP is a way for computers to understand, analyze, and derive meaning from human language in a smart and useful way. We introduce the following AI methods applied to NLP.

Recurrent neural networks (RNN) [37] make full use of sequential information. As we all know, the inputs and outputs of traditional neural networks are independent. In practice, it must predict the words before a sentence. The so-called RNN is a recurrent network because it performs the same task for every element of a sequence, with the output being dependent on the previous computations. There are many types of improved RNN models that have been proposed to solve some of the shortcomings of the original RNN model. Bidirectional RNN [38] is based on the principle that the output may not depend only on the previous elements in the sequence but also on the future elements. Deep bidirectional RNN [39] is similar to Bidirectional RNN but improved by adding multiple layers per time step. Long short-term memory (LSTM) [40] uses the same mechanism to decide what to keep in and what to erase from memory that is used in RNNs.

Recursive neural network [41] is another deep neural network created by applying the same set of weights recursively over a structure in order to produce a structured prediction over the input by transferring a given structure in topological order. Dependency neural network (DCNN) [53] is a method proposed to capture long-distance dependencies. DCNN consists of a convolutional layer built on top of an LSTM model. Dynamic k-max pooling neural network [54] is another type of network that uses a non-linear max pooling subsampling operator to return the maximum of a set of values. This network outputs k maximum values in the sequence and optimizes select k values by other functions. Multi-column CNN [55] shares the same word embedding and multiple columns of convolutional neural networks. Ranking CNN [56] takes the relation classification task using a convolutional neural network that performs classification by ranking. Context-dependent CNN [57] consists of two components: a convolutional sentence model that summarizes the meaning of the source sentence and the target phrase and a matching model that compares the two representations with a multilayer perceptron. Sample vendors include Basis Technology, Coveo, Expert System, Indico, Knime, Lexalytics, Linguamatics, Mindbreeze, Sinequa, Stratifyd, and Synapsify.

## 2.9 Visual Recognition

Deep learning has been shown to be one of the best solutions for computer vision. A large number of methods have been developed to improve the performance of traditional deep-learning algorithms. In general, these methods can be divided into three categories: convolutional neural networks, autoencoders, and sparse and restricted Boltzmann machines. In this paper, we focus on searching convolutional neural network models.

The pipeline of the traditional convolutional neural network architecture consists of three main sets of layers: convolutional, pooling, and fully connected layers. Different layers play different roles in the classification. Convolutional layers are used to convolve the image to generate feature maps. The main advantages of convolutional layers are that the weight-sharing mechanism reduces the number of parameters and local connectivity learns the relationships between the neighbor pixels. In addition, it is invariant to the location of the objects in the image.

The pooling layers are usually used after the convolutional layers to reduce the dimensions of the feature maps and to adjust the parameters. Average pooling and max pooling are used in most cases. Following the last pooling layers, the fully connected layers are used to convert the two-dimensional feature maps into one-dimensional feature vectors. Several state-of-the-art convolutional neural network models are reviewed below.

Convolutional neural networks (CNN) [43] are similar to traditional neural networks (NN). They are made up of neurons that have learnable weights and biases. The main difference between CNN and NN is the number of layers. CNN use several layers of convolutions with nonlinear activation functions applied to the results.

AlexNet [42] contains eight layers. The first five layers are the convolutional layers, and the following three layers are the fully connected layers. Compared to CNN [43], AlexNet has advantages such as data augmentation, dropout [44], ReLU, local response normalization, and overlapping pooling.

The main contribution of VGGNet [45] is to increase the network depth using very small convolution filters. The total number of layers in VGGNet is 16–19. However, the use of max-pooling layers results in a loss of accurate spatial information.

Szegedy et al. [46] contributed to improving the use of computing resources inside a network. The GoogLeNet method increases the width and depth of the network while keeping the computational budget constant. Based on Arora et al.'s research [47], the layer-by-layer construction can analyze the correlation statistics of the last layer and then combine them into groups. One of the main advantages of GoogLeNet is that it allows the number of layers at each stage to be increased without an uncontrolled blow-up in the computational complexity. Another benefit is that this network is 2–3 times faster than similarly performing networks. However, it is complex to configure the design of this network.

There is a trend for deeper layers to result in better network performance. However, with increasing network depth, the training accuracy becomes saturated and then rapidly degrades. He et al. [48] solved this problem using a deep residual learning framework. Additional "shortcut connections" are added to feedforward neural networks because short connections add neither extra parameters nor computational complexity.

## 3. Brain Intelligence (BI)

There are many approaches [58-61] proposed to solve the limitations of recent AI. However, these models are simply extended from the current AI models. This paper introduces the following items for explaining the concept of BI, which is different from artificial intelligence, but extends upon current artificial intelligence.

The BI intelligent learning model fuses the benefits of artificial life (AL) and AI. Currently, the mainstream research on deep learning is a method of learning expressions extracted from essential information of observational data by a deep neural network with a large number of layers. However, research on multitask learning that learns multiple tasks at the same time and transition studies that divert learning results for a certain task to other tasks is still insufficient. For this reason, AI models based on unsupervised learning and shallow neural networks will become trends in future. In this paper, we will combine various regional AI methods using a particular rule, especially unsupervised learning methods. It is essential to develop a new intelligent learning model with a small database and the ability to understand concepts. Therefore, we propose a Brain Intelligence model with memory and idea function in Figure 3. The BI model network combines artificial life technology and artificial intelligence technology with memory function.

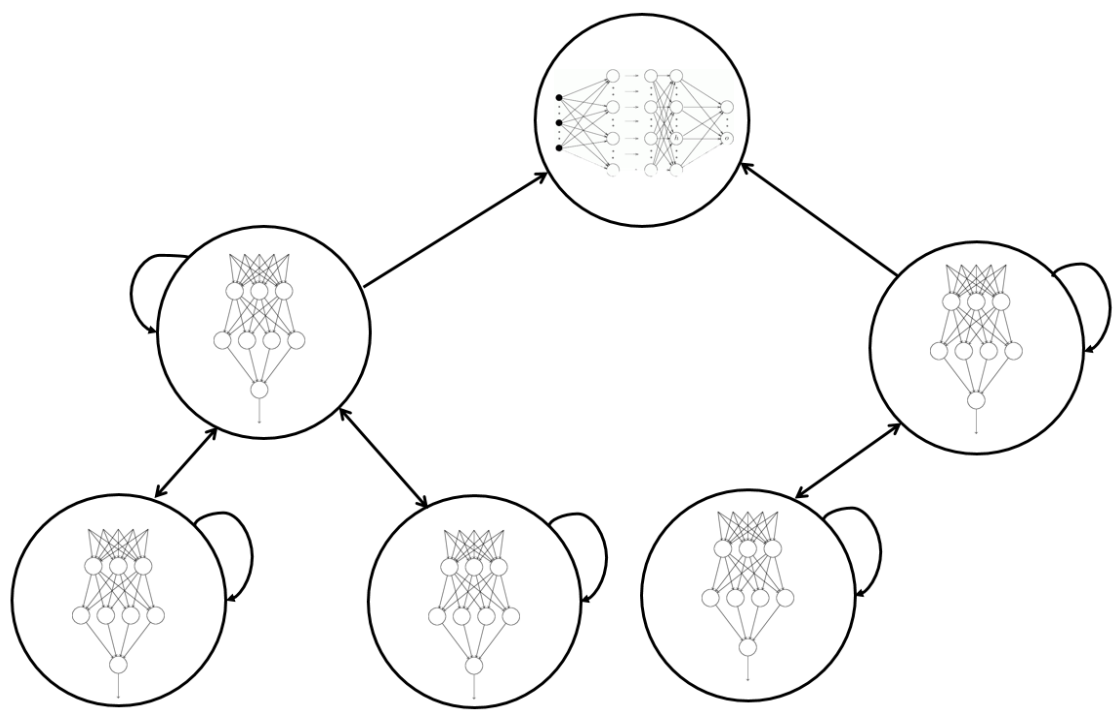

Figure 3. The concept of the BI model network. Different neural networks are connected by artificial life-based network, which can share the parameters, trained results, and structures to parents and sons.

Research on current AI mainly focuses on individual areas such as dialogue comprehension, visual recognition, and auditory discrimination and so on. Research on whole-brain functions is still insufficient. For example, there are few studies on perceptual understanding models and self-thinking models. Therefore, in this research, we will clarify the function and mechanisms of the whole brain and make efforts to realize it as artificial intelligence. BI network is consisted by many simple sub-networks. The parameters of each sub-networks is updated by S-system [62], which can modify the sub-networks by reproduction, selection, and mutation.

Different from NeuroEvolution of Augmenting Topologies (NEAT) [63, 64], the proposed BI mode network does not just use the neural network structure and parameter optimization mechanism, it improves the structure of current AI models using S-system. hyperNEAT [65], a type of A-life based NN, which uses the Compositional Pattern Producing Network (CPPN) for pattern generation and uses NEAT for parameters optimization. hyperNEAT cannot overcome the drawbacks of the NEAT network. Other gene-based models, such as Gene Regulatory Network (GRN) [66] and Evolving Reaction Network (ERN) [67], are also studied by some researchers. These methods are inspired by biological characteristics, which do not take into account the usage of all the brain's function. Cognitive Computing (CC) [68] is proposed a new model from the view of human cognitive functions. The BI model network is investigated from an engineering point of view, in the future, we will develop a super-intelligent brain function model that intends to discover problems itself and autonomously enhance its abilities.

## 4. Conclusion

In this paper, we have presented state-of-the-art artificial intelligence tools for individual application areas, such as natural language processing and visual recognition. The main contributions of this work are as follows. First, this is an overview of current deep learning methods. We have summarized the nine potential applications in detail. Second, this paper puts together all the problems of recent AI models, which will direct future work for researchers. Third, in this paper, we first proposed the brain intelligence model, which is a model fusing artificial intelligence and artificial life. AL models, such as the S-system, have the benefits of an association function, which is different from generative adversarial networks (GAN), for building big data within a life evolution process. It is foreseeable that the BI model can solve the issues of the frame problem, the association function problem, the symbol grounding problem, and the mental/physical problem.

**Acknowledgements**

This work was supported by Leading Initiative for Excellent Young Researcher (LEADER) of Ministry of Education, Culture, Sports, Science and Technology-Japan (16809746), Grants-in-Aid for Scientific Research of JSPS (17K14694), Research Fund of Chinese Academy of Sciences (No.MGE2015KG02), Research Fund of State Key Laboratory of Marine Geology in Tongji University (MGK1608), Research Fund of State Key Laboratory of Ocean Engineering in Shanghai Jiaotong University (1510), Research Fund of The Telecommunications Advancement Foundation, and Fundamental Research Developing Association for Shipbuilding and Offshore.

**References**

[1] Siri, https://en.wikipedia.org/wiki/Siri (Accessed on 2017/4/20).
[2] AlphaGo, https://deepmind.com/research/alphago/ (Accessed on 2017/4/20).
[3] IBM Watson, https://www.ibm.com/watson/ (Accessed on 2017/4/20).
[4] Microsoft Translator Speech API, https://www.microsoft.com/en-us/translator/speech.aspx (Accessed on 2017/4/20).
[5] Amazon Prime Air, https://www.amazon.com/Amazon-Prime-Air/b?node=8037720011 (Accessed on 2017/4/20).
[6] Y. Taigman, M. Yang, M. Ranzato, L. Wolf, "DeepFace: Closing the Gap to Human-Level Performance in Face Verification," IEEE International Conference on Computer Vision and Pattern Recognition (CVPR2014), pp.1-8, 2014.
[7] Stanford Artificial Intelligence Laboratory, http://ai.stanford.edu/ (Accessed on 2017/4/20).

[8] MIT BigDog, https://slice.mit.edu/big-dog/ (Accessed on 2017/4/20).

[9] The 4th Science and Technology Basic Plan of Japan, http://www8.cao.go.jp/cstp/english/basic/ (Accessed on 2017/4/20).

[10] AI EXPO, http://www.ai-expo.jp/en/ (Accessed on 2017/4/20).

[11] 2017 Will Be the Year of AI, http://fortune.com/2016/12/30/the-year-of-artificial-intelligence/ (Accessed on 2017/4/20).

[12] Y. LeCun, Y. Bengio, G. Hinton, "Deep learning," Nature, vol.521, no.7553, pp.436-444, 2015.

[13] Y. Bengio, R. Ducharme, P. Vincent, C. Janvin, "A neural probabilistic language model," Journal of Machine Learning Research, vol.3, pp.1137-1155, 2003.

[14] A. Mnih, G. Hinton, "Three new graphical models for statistical language modelling," In Proc of ICML07, pp.641-648, 2007.

[15] T. Mikolov, M. Karafiat, L. Burget, J. Cernocky, S. Khudanpur, "Recurrent neural network based language model," In Proc of Interspeech10, pp.1045-1048, 2010.

[16] I. Sutskever, O. Vinyals, Q. Le, "Sequence to sequence learning with neural networks," In Advances in Neural Information Processing Systems, pp.3104-3112, 2014.

[17] H. Mei, M. Bansal, M. Walter, "What to talk about and how? Selective generation using LSTMs with coarse-to-fine alignment," In NAACL-HLT, pp.1-11, 2016.

[18] M. Luong, Q. Le, I. Sutskever, O. Vinyals, L. Kaiser, "Multitask sequence to sequence learning," In Proc ICLR, pp.1-10, 2016.

[19] H. Bourlard, M. Morgan, "Connnectionist speech recognition: a hybrid approach," Kluwer Academic Publishers, 1994.

[20] A. Graves, A. Mohamed, G. Hinton, "Speech recognition with deep recurrent neural networks," in ICASSP2013, pp.1-5, 2013.

[21] B. Wiederhold, G. Riva, M. Wiederhold, "Virtual reality in healthcare: medical simulation and experiential interface," Annual Review of Cyber Therapy and Telemedicine, vol.13, 239 pages, 2015.

[22] G. Bartsch, A. Mitra, S. Mitra, A. Almal, K. Steven, D. Skinner, D. Fry, P. Lenehan, W. Worzel, R. Cote, "Use of artificial intelligence and machine learning algorithms with gene expression profiling to predict recurrent nonmuscle invasive urothelial carcinoma of the bladder," The Journal of Urology, vol.195, pp.493-498, 2016.

[23] N. Labonnote, K. Hoyland, "Smart home technologies that support independent living: challenges and opportunities for the building industry – a systematic mapping study," Intelligent Buildings International, vol.29, no.1, pp.40-63, 2017.

[24] S. Chetlur, C. Woolley, P. Vandermersch, J. Cohen, J. Tran, B. Catanzaro, E. Shelhamer, "Cudnn: efficient primitives for deep learning," pp.1-10, arXiv:1410.0759, 2014.

[25] A. Coates, B. Huval, T. Wang, D. Wu, B. Catanzaro, N. Andrew, "Deep learning with cots hpc systems," In Proc of the 30th International Conference on Machine Learning, pp.1337-1345, 2013.

[26] G. Lacey, G. Taylor, S. Areibi, "Deep learning on FPGAs: past, present, and future," pp.1-8, arXiv: 1602.04283, 2016.

[27]W. Lin, S. Lin, T. Yang, "Integrated business prestige and artificial intelligence for corporate decision making in dynamic environments," Cybernetics and Systems, DOI: 10.1080/01969722.2017.1284533, pp.1-22, 2017.

[28]A. Ratnaparkhi, E. Pilli, R. Joshi, "Survey of scaling platforms for deep neural networks," In Proc of International Conference on Emerging Trends in Communication Technologies, pp.1-6, 2016.

[29]R. Raina, A. Madhavan, A. Ng, "Large-scale deep unsupervised learning using graphics processors," In Proc of 26$^{th}$ Annual International Conference on Machine Learning, pp.873-880, 2009.

[30]B. Catanzaro, "Deep learning with COTS HPC systems," In Proc of the 30$^{th}$ International Conference on Machine Learning, pp.1337-1345, 2013.

[31]J. Dean, G. Corrado, R. Monga, K. Chen, M. Devin, Q. Le, M. Mao, M. Ranzato, A. Senior, P. Tucker, K. Yang, A. Ng, "Large scale distributed deep networks," In Proc of Advances in Neural Information Processing Systems, pp.1223-1231, 2012.

[32]T. Chilimbi, Y. Suzue, J. Apacible, K. Kalyanaraman, "Project adam: Building an efficient and scalable deep learning training system," In Proc of 11$^{th}$ USENIX Symposium on Operating Systems Design and Implementation, pp.571-582, 2014.

[33]Qualcomm Zeroth, https://www.qualcomm.com/invention/cognitive-technologies/zeroth (Accessed on 2017/4/27).

[34]P. Merolla, J. Arthur, R. Alvarez-lcaza, A. Cassidy, J. Sawada, F. Akopyan, B. Jackson, N. Imam, C. Guo, Y. Nakamura, B. Brezzo, I. Vo, S. Esser, R. Appuswamy, B. Taba, A. Amir, M. Flickner, W. Risk, R. Manohar, D. Modha, "A million spiking-neuron integrated circuit with a scalable communication network and interface," Science, vol.345, no.6197, pp.668-673, 2014.

[35]M. Khan, D. Lester, L. Plana, A. Rast, X. Jin, E. Painkras, S. Furber, "SpiiNNaker: Mapping neural networks onto a massively-parallel chip multiprocessor," In Proc of IEEE International Joint Conference on Neural Networks, pp.2849-2856, 2008.

[36]M. Lacity, L. Willcocks, "A new approach to automating services," MIT Sloan Management Review, vol.2016, pp.1-16, 2016.

[37]T Mikolov, M. Karafiat, L. Burget, J. Cernocky, S. Khudanpur, "Recurrent neural network based language model," In Proc of Interspeech2010, pp.1045-1048, 2010.

[38]M. Schuster, K. Paliwal, "Bidirectional recurrent neural networks," IEEE Transactions on Signal Processing, vol.45, no.11, pp.2673-2681, 1997.

[39]A. Graves, N. Jaitly, A. Mohamed, "Hybrid speech recognition with deep bidirectional LSTM," In Proc of IEEE Workshop on Automatic Speech Recognition and Understanding, pp.1-4, 2013.

[40]A. Graves, J. Schmidhuber, "Framewise phoneme classification with bidirectional LSTM and other neural network architecutres," Neural Networks, vol.18, no.5-6, pp.602-610, 2005.

[41]A. Mishra, V. Desai, "Drought forecasting using feed-forward recursive neural network," Ecological Modelling, vol.198, no.1-2, pp.127-138, 2006.

[42]A. Karpathy, G. Toderici, S. Shetty, T. Leung, R. Sukthankar, and F. Li, "Large-scale video classification with convolutional neural networks," In Proc of IEEE Conference on Computer Vision and Pattern Recognition, pp.1725-1732, 2014.

[43]Y. LeCun, B. Boser, J. Denker, D. Henderson, R. Howard, W. Hubbard, and L. Jackel, "Backpropagation applied to handwritten zip code recognition," Neural Computation, vol.1, no.4, pp.541-551, 1989.

[44] R. Bell, and Y. Koren, "Lessons from the Netflix prize challenge," ACM SIGKDD Explorations Newsletter, vol.9, no.2, pp.75-79, 2007.

[45]K. Simonyan, and A. Zisserman, "Very deep convolutional networks for large-scale image recognition," In Proc of IEEE ICLR2015, pp.1-14, 2015.

[46]C. Szegedy, W. Liu, Y. Jia, P. Sermanet, S. Reed, D. Anguelov, D. Erhan, V. Vanhoucke, A. Rabinovich, "Going deeper with convolutions," In Proc of IEEE Conference on Computer Vision and Pattern Recognition, pp.1-12, 2015.

[47]S. Arora, A. Bhaskara, R. Ge, and T. Ma, "Provable bounds for learning some deep representations," arXiv:abs/1310.6343, 2013.

[48]K. He, X. Zhang, S. Ren, J. Sun, "Deep residual learning for image recognition," In Proc. Of IEEE Conference on Computer Vision and Pattern Recognition, pp.1-12, 2016.

[49]M. Chen, Y. Ma, Y. Li, D. Wu, Y. Zhang, "Wearable 2.0: Enabling Human-Cloud Integration in Next Generation Healthcare Systems," IEEE Communications Magazine, vol. 54, no. 12, pp. 3-9, 2017.

[50] J. Song, Y. Zhang, "TOLA: Topic-oriented learning assistance based on cyber-physical system and big data," Future Generation Computer Systems, DOI:10.1016/j.future.2016.05.040, 2016.

[51]Y. Zhang, "Grorec: A group-centric intelligent recommender system integrating social, mobile and big data technologies," IEEE Transactions on Services Computing, vol. 9, no. 5, pp. 786-795, 2016.

[52]Q. Liu, Y. Ma, M. Alhussein, Y. Zhang, L. Peng, "Green data center with IoT sensing and cloud-assisted smart temperature controlling system," Computer Networks, Vol. 101, pp. 104-112, June 2016.

[53]D. Chen, C. Manning, "A fast and accurate dependency parser using neural networks," In Proc of Empirical Methods in Natural Language Processing, pp.740-750, 2014.

[54]N. Kalchbrenner, E. Grefenstette, P. Blunsom, "A convolutional neural network for modelling sentences," In Proc of Annual Meeting of the Association for Computational Linguistics, pp.655-665, 2014.

[55]D. Ciresan, U. Meier, J. Masci, J. Schmidhuber, "Multi-column deep neural network for traffic sign classification," Neural Networks, vol.32, pp.333-338, 2012.

[56]C. Santos, B. Xiang, B. Zhou, "Classifying relations by ranking with convolutional neural

networks," In Proc of Annual Meeting of the Association for Computational Linguistics, pp.626-634, 2015.

[57] B. Hu, Z. Tu, Z. Lu, Q. Chen, "Context-dependent translation selection using convolutional neural network," In Proc of Annual Meeting of the Association for Computational Linguistics, pp.536-541, 2015.

[58] Y. Li, H. Lu, J. Li, X. Li, Y. Li, S. Serikawa, "Underwater image de-scattering and classification by deep neural network," Computers & Electrical Engineering, vol.54, pp.68-77, 2016.

[59] H. Lu, B. Li, J. Zhu, Y. Li, Y. Li, X. Xu, L. He, X. Li, J. Li, S. Serikawa, "Wound intensity correction and segmentation with convolutional neural networks," Concurrency and Computation: Practice and Experience, vol.29, no.6, pp.1-8, 2017.

[60] H. Lu, Y. Li, T. Uemura, Z. Ge, X. Xu, L. He, S. Serikawa, H. Kim, "FDCNet: filtering deep convolutional network for marine organism classification," Multimedia Tools and Applications, pp.1-14, 2017.

[61] H. Lu, Y. Li, L. Zhang, S. Serikawa, "Contrast enhancement for images in turbid water," Journal of the Optical Society of America, vol.32, no.5, pp.886-893, 2015.

[62] S. Serikawa, T. Shimomura, "Proposal of a system of function-discovery using a bug type of artificial life," Transactions of IEE Japan, vol.118-C, no.2, pp.170-179, 1998.

[63] K. Stanley, R. Miikkulainen, "Evolving neural networks through augmenting topologies," Evolutionary Computation, vol.10, no.2, pp.99-127, 2002.

[64] J. Schrum, R. Miikkulainen, "Evolving multimodal behavior with modular neural networks in Ms. Pac-Man," in Proc of the Genetic and Evolutionary Computation Conference, pp.325-332, 2014.

[65] K. Stanley, D. Ambrosio, J. Gauci, "A hypercube-based encoding for evolving large-scale neural networks," Artificial Life, vol.15, no.2, pp.185-212, 2009.

[66] F. Emmert-Streib, M. Dehmer, B. Haibe-Kains, "Gene regulatory networks and their applications: understanding biological and medical problems in terms of networks," Frontiers in Cell and Developmental Biology, vol.2, no.38, pp.1-7, 2014.

[67] H. Dinh, M. Aubert, N. Noman, T. Fujii, Y. Rondelez, H. Iba, "An effective method for evolving reaction networks in synthetic biochemical systems," IEEE Transactions on Evolutionary Computation, vol.19, no.3, pp.374-386, 2015.

[68] K. Hwang, M. Chen, "Big-Data analytics for cloud, IoT and cognitive computing," Wiley Press, 432 pages, 2017.